\documentclass{article} 
\usepackage{iclr2018_conference,times}
\usepackage{hyperref}
\usepackage{url}
\usepackage{graphicx}
\usepackage{comment}
\usepackage{amsmath}
\usepackage{todonotes}
\usepackage{changebar}
\usepackage{amssymb}
\usepackage{pifont}

\title{Hierarchical Block Sparse Neural Networks}

\author{Dharma Teja Vooturi \\
Center for Security Theory and Algorithmic Research\\
International Institute of Information Technology - Hyderabad\\
Hyderabad, India \\
\texttt{\{dharmateja.vooturi\}@research.iiit.ac.in} \\
\AND
Dheevatsa Mudigere\thanks{Author is now affiliated with Facebook.} , Sashikanth Avancha \\
Parallel Computing Lab \\
Intel Labs, India \\
\texttt{\{sasikanth.avancha\}@intel.com} \\
}

\iclrfinalcopy 

\begin{document}

\maketitle

\begin{abstract}
Sparse deep neural networks(DNNs) are efficient in both memory and compute when compared to dense DNNs. But due to irregularity in computation of sparse DNNs, their efficiencies are much lower than that of dense DNNs on regular parallel hardware such as TPU. This inefficiency leads to poor/no performance benefits for sparse DNNs. Performance issue for sparse DNNs can be alleviated by bringing structure to the sparsity and leveraging it for improving runtime efficiency. But such structural constraints often lead to  suboptimal accuracies. In this work, we jointly address both accuracy and performance of sparse DNNs using our proposed class of sparse neural networks called \textbf{HBsNN} (\textbf{H}ierarchical \textbf{B}lock \textbf{s}parse \textbf{N}eural \textbf{N}etworks). For a given sparsity, HBsNN models achieve better runtime performance than unstructured sparse models and better accuracy than highly structured sparse models. 

\end{abstract}

\section{Introduction}
Deep learning is playing a pivotal role in advancing artificial intelligence. Modern day deep neural networks(DNNs) use millions of parameters to perform well on the task at hand. For instance, Convolutional Neural Networks(CNNs) such as Resnet-50 \citet{he2016identity} and Inception-v3 \cite{szegedy2016rethinking} use $\sim$25 and $\sim$23.8 million parameters respectively to achieve state of the art accuracies for image classification task. In general, deep learning methods are data savvy and with increase in data, models with more complexity i.e, more number of parameters are required to achieve better accuracies. This results in an increase in both memory footprint, and compute of the model. 

In neural networks, each parameter is equally important before the training begins. As the training progresses, importance of these parameters vary. One can prune away least important parameters during or after the training process with minimal/no loss to the model accuracy. Pruning parameters leads to two benefits: 1) Memory footprint of the model is reduced as we need not store pruned parameters. 2) Computational complexity is decreased as we need not do multiplications involved with pruned parameters. Thus models which are both memory and compute efficient can be generated using pruning techniques. Early studies by  \cite{Cun90optimalbrain,Hassibi93secondorder}, have shown the efficacy of pruning technique in reducing the model complexity. More recently, pruning techniques were successfully applied on many classes of neural networks: On Convolutional Neural Networks(CNNs), \cite{han2015learning} was able to generate sparse CNNs by pruning parameters from a pretrained dense CNNs and follow it by finetuning. On recurrent neural networks(RNNs), \cite{DBLP:journals/corr/NarangDSE17} was able to generate sparse RNNs by pruning away parameters at regular intervals during the training process. And also, pruning serves as an effective technique for model compression and can be used alongside with other model compression techniques.

Most common way of pruning a neural network is fine grained pruning, where pruning is performed at the level of individual element and the sparsity obtained due to it is unstructured. For a given neural network, if K\% of the model parameters are uniformly pruned across all layers, the computational complexity of the model reduces by a factor of 100/(100-K). For example, pruning half of the parameters in a model decreases the computational complexity by 2x. But for fine grained pruning, the runtime benefit obtained by decrease in computational complexity is far from ideal on regular parallel hardware such as TPU. Specialized sparse accelerators \cite{Han:2016:EEI:3007787.3001163}, \cite{DBLP:journals/corr/abs-1708-04485} have to be built to cash in benefits of reduced computational complexity from fine grained pruning.

To deal with runtime performance issue of unstructured sparse neural networks, researchers have resorted to pruning parameters in a more structured way and leverage the structure for runtime performance. Towards that end, \cite{DBLP:journals/corr/abs-1711-02782} have performed block pruning in Recurrent Neural Networks(RNNs), and \citep{Wen:2016:LSS:3157096.3157329}, have performed filter pruning. But the common observation is that for a given sparisty, the model obtained by these highly structured pruning(block, filter) have less accuracy than fine grained pruning.

Fine grained pruning lacks runtime performance, and highly structured pruning lacks model accuracy. But we would like our sparse networks to have both accuracy and performance. In this work, we arrive at such sparse models using our proposed class of sparse neural networks called  \textbf{HBsNN} (\textbf{H}ierarchical  \textbf{B}lock \textbf{s}parse \textbf{N}eural \textbf{N}etworks). The main idea is to have multiple hierarchical structural components which caters for both accuracy and performance. In Table \ref{table:summary}, we compare HBsNN on three important metrics with other sparsity types.

\begin{table}[h!]
\begin{center}
\begin{tabular}{| p{10em} | p{5em} | p{8em} | p{8em}|}
\hline
 & \multicolumn{3}{|c|}{Sparsity type} \\ \hline
Metrics & Unstructured sparsity & Highly Structured  sparsity & Hierarchical block sparsity(HBS)\\ \hline
Low Memory foot print & \ding{51} & \ding{51} & \ding{51} \\ \hline
Model accuracy & \ding{51} & \ding{55} & \ding{51} \\ \hline
Performance &  \ding{55} &  \ding{51} & \ding{51} \\ \hline
\end{tabular}
\end{center}
\caption{Fixing sparsity in Neural Networks.}
\label{table:summary}
\end{table}

\textbf{Contributions}
\begin{itemize}
\item Proposed a class of sparse neural networks called HBsNN(Hierarchical Block Sparse Neural Networks), which caters for both accuracy and performance.
\item Designed a performance model for the compute in HBsNN.
\end{itemize}

\section{HBsNN}
\subsection{Motivation}
Importance of a parameter in a neural network is strongly correlated with it's magnitude. When we perform highly structured pruning like block sparse, we lose significant number of high magnitude parameters due to the imposed structural constraints. Row 1 in Table \ref{table:topk}, shows the percentage of top \{10,20,30,40,50\}\% elements retained after pruning 50\% of elements in a block sparse manner using 32x1 block size on a pretrained Resnet-v2-50 model. One can see that $~$24\% of the top 10\% elements are pruned out. It has been found empirically that high magnitude parameters play a significant role in generating sparse models with good model accuracies. One simple way to retain high magnitude weights is to bring fluidity to the sparsity structure. In Table \ref{table:topk}, one can see that for a given sparsity of 50\%, incorporating multiple levels of structure leads to improved top-* percentages. Based on this observation, we propose a class of sparse neural networks called \textbf{H}ierarchical \textbf{B}lock \textbf{s}parse \textbf{N}eural \textbf{N}etworks(HBsNN) which are more fluid and can retain high magnitude parameters. Sparse models obtained by fine grained pruning or block sparsity are a subset of HBsNN. 

\begin{table}[h!]
\centering
\begin{tabular}{|c|c|c|c|c|c|}
\hline
(block-heightx1)/sparsity	& top-10 & top-20 & top-30 & top-40 & top-50  \\ \hline
32/50         & 76.04 & 69.81 & 65.80 & 62.82 & 60.42 \\ \hline
(32,16)/(75, 75) & 80.25 & 72.95 & 68.16 & 64.63 & 61.80 \\ \hline
(32,16,8)/(75, 87.5, 87.5) & 85.00 & 76.56 & 70.91 & 66.77 & 63.40 \\ \hline
(32,16,8,4)/(75, 87.5, 93.75, 93.75) & 89.83 & 80.06 & 73.49 & 68.63 & 64.75 \\ \hline
(32,16,8,4,1)/(75, 87.5, 93.75, 96.875, 96.875) & 100 & 90.71 & 79.23 & 72.03 & 66.84 \\ \hline
\end{tabular}
\caption{HBS configuration vs Retained percentage on ResNet-v2-50 model.}
\label{table:topk}
\end{table}

\subsection{Description}
\begin{figure}[h!]
\includegraphics[width=\textwidth]{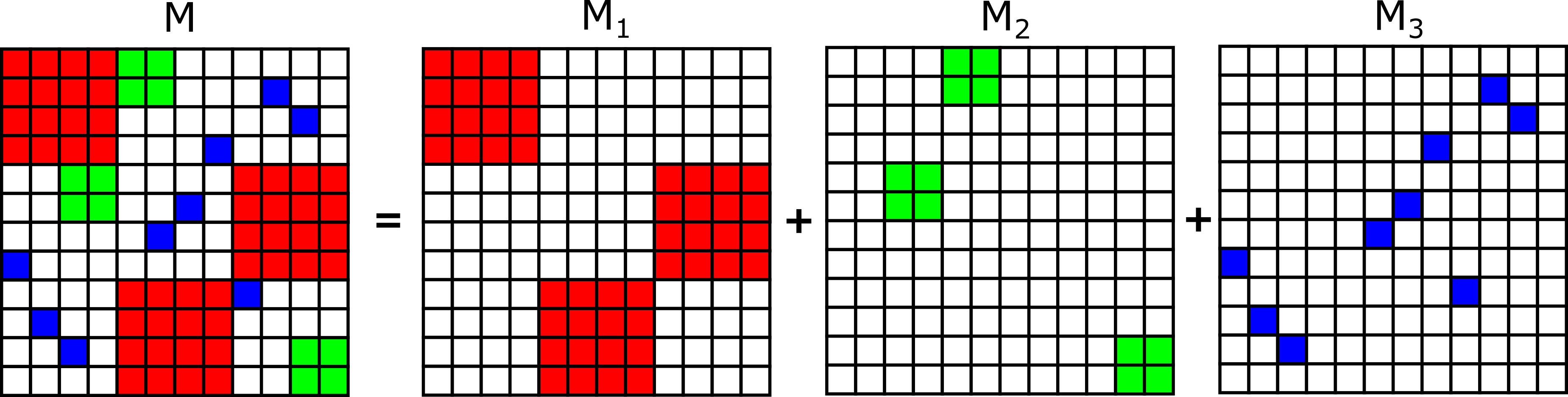}
\caption{Hierarchical block sparse(HBS) configuration with 3 levels[(4x4),(2x2) and (1x1)]. }
\label{fig:mlbsnn-example}
\end{figure}

In HBsNN (Hierarchical Block Sparse Neural Networks), sparse parameter matrix $M$ of a given layer is composed of multiple sparse parameter matrices i.e, $M = M_1 + ...+ M_N$, where each $M_i$ is a block sparse matrix with different block dimensions. As it is suboptimal to split the value of a non-zero in $M$ across many matrix levels, a non-zero element in M is contributed by only one $M_i$ i.e, $\forall_{j,k \in N} M_j .* M_k = 0$. 
Apart from this, matrices have to satisfy hierarchical structure, where dimensions of block in $M_{i+1}$ should divide dimensions of block in $M_{i}$ i.e, $Dim(M_{i})\%Dim(M_{i+1}) = 0$.
In Figure \ref{fig:mlbsnn-example}, we have a 3 level configuration with block dimensions 4x4,2x2 and 1x1 respectively.

\subsection{Pruning methodology}
\begin{figure}[h!]
\centering
\includegraphics[scale=0.25]{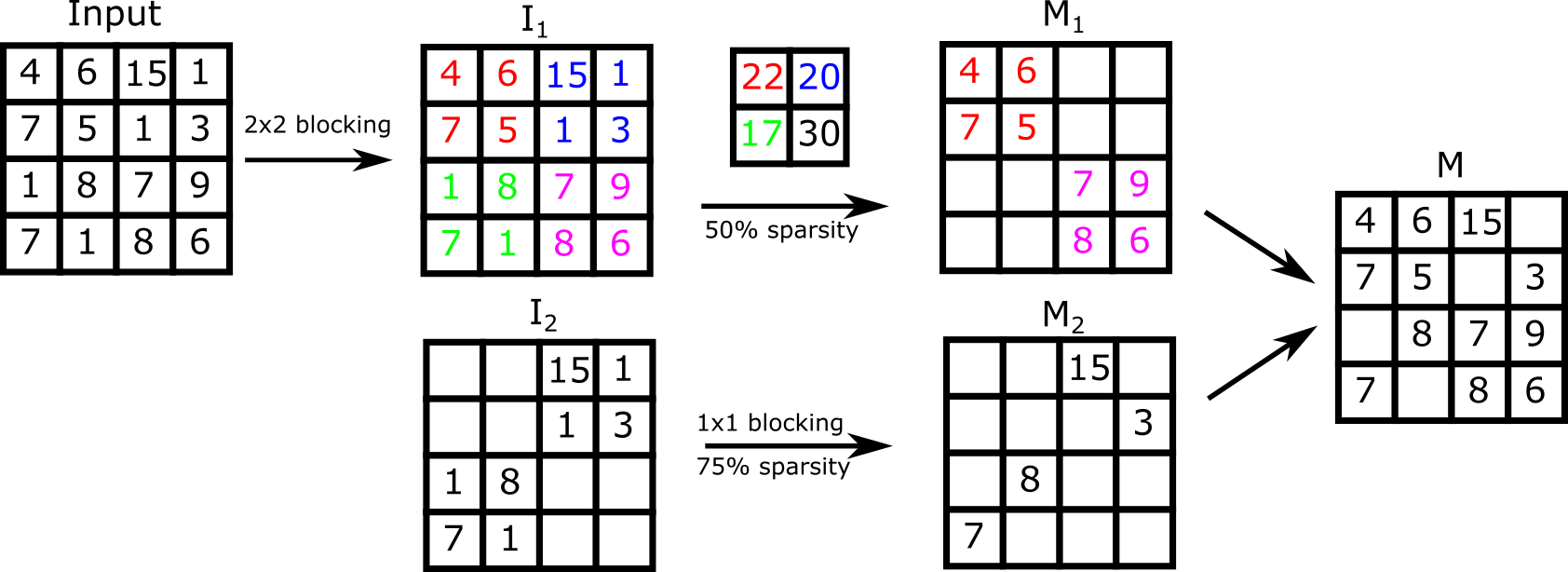}
\caption{2-Level block sparse generation : Block sizes=[(2x2),(1x1)] sparsities=[50,75]}
\label{fig:mlbsnn-pruning}
\end{figure}

For a given matrix $I$, block dimensions $(bh,bw)$ and sparsity $sp$, block sparsity is generated by dividing the matrix $I$ into a grid, where each grid element is of size $(bh,bw)$. Each grid element is then given a rank using the absolute summation values of that grid block. We then sort these values and prune away $sp$ \% of blocks to generate a block sparse matrix. In case of hierarchical block sparsity with $N$ levels, block sizes $BS = [(bh_1,bw_1), \dots (bh_N,bw_N)]$ and sparsities $SP = {sp_1, \dots sp_N}$ are provided for all the levels. Let $I_k$ and $M_k$ be the input and output matrices at level $k$. In level $k$, we perform a block sparse pruning with block size $(bh_k,bw_k)$ and sparsity $sp_k$ to generate $M_k$. We then generate input to layer $k+1$ by removing elements of $M_k$ from $I_k$ i.e, $I_{k+1} = I_k - M_k$. Figure \ref{fig:mlbsnn-pruning}, shows an example of 2 level HBS pruning on 4x4 matrix with BS=[(2,2),(1,1)] and SP=[50,75]. In case of networks where parameters of a layer are arranged in more than two dimensions, block dimensions correspond to outer two dimensions. For example, in case of CNNs, blocking is performed on ofm(output feature maps) and ifm(input feautre maps).
  

\subsection{Performance model}
In this section, we describe a performance model for evaluating performance of a layer in HBsNN. For a given HBS configuration, let $F_{dense}$  and $F_{sparse}$ be the amount of compute for dense and sparse operations respectively. As a layer in HBsNN has multiple levels($L_1,\dots,L_N$), $F_{sparse} = \sum_{i=1}^{i=N} F_{sparse}^{L_i}$, where $F_{sparse}^{L_i}$ is the amount of compute in $i^{th}$ level of the layer. Due to irregularity in sparse computation, it is not always possible to realize the ideal speed up i.e, $F_{dense}/F_{sparse}$. The achievable speedup depends primarily on two factors: 1)Amount of sparsity and 2)Dimensions of blocks. We quantify the sub-optimal speedup with an irregular factor function $irf(sparsity, blockDimensions)$ parameterized by those two factors. By taking these factors into effect, the cost of dense($C_{dense}$), and the cost of sparse neural network($C_{sparse}$) are defined according to equation 1 and 2. Achievable speedup can then be defined as $C_{dense}/C_{sparse}$. $irf(\dots)$ function varies from a system to system and has to be obtained by running micro benchmarks on that system. But on a regular parallel hardware such as TPU, $irf$ function is inversely proportional to block size. So, inorder to maximize performance, one needs to maximize sparsity for levels with smaller block sizes, and minimize sparsity for levels with larger block sizes.

\noindent\begin{minipage}{.25\linewidth}
\begin{equation}
C_{dense} = F_{dense}
\end{equation}
\end{minipage}
\noindent\begin{minipage}{.75\linewidth}
\begin{equation}
C_{sparse} = \sum_{i=1}^{i=N}  \dfrac{ F_{sparse}^{L_i}}{irf(Sparsity(L_{i}), BlockDims(L_{i}))}
\end{equation}
\end{minipage}
\begin{minipage}{.3\linewidth}
\begin{equation}
SpeedUp = \dfrac{C_{dense}}{C_{sparse}}
\end{equation}
\end{minipage}

\section{Results}
\subsection{ResNet-v2-50/Imagenet}
We took a pretrained Renset-v2-50 model with top-1 and top-5 accuracy of 76.13\%  and 92.86\% respectively and then generated sparse models from it using prune and retrain methodology from \cite{han2015learning}. Except for the first convolution layer and the last fc layer, all layers are pruned. The pruned model is then trained for 18 epochs with the same set of hyper parameters as that of the pretrained model. The initial learning rate for training is set to ${1/100}^{th}$ of the base learning rate used for pretrained model. A step based learning rate decay is followed, where learning rate is decreased by a factor of 10 and 100 respectively at $9^{th}$ and $14^{th}$ epoch respectively. 

\textbf{Varying sparsity} : In this experiment, we would like to study how accuracy varies with respect to sparsity. We vary sparsity from 50 to 87.50 with block size set to 1x1. From Table \ref{table:resnet50-vs}, we can see that accuracy decreases with sparsity and the rate at which it decreases is exponential. This is due to the fact that more number of elements are pruned away with increase in sparsity and this reduces the model capacity.

\begin{table}[h!]
\centering
\begin{tabular}{|c|c|c|}
\hline
Sparsity  & Top-1 Accuracy & Top-5 Accuracy\\ \hline
50 & 76.42 (+0.29) & 93.03 (+0.17) \\ \hline
75 & 75.12 (-1.01) & 92.34 (-0.52) \\ \hline
87.50 &71.58 (-4.55) & 90.58 (-2.28) \\ \hline
\end{tabular}
\caption{Varying sparsity with block size set to 1x1.}
\label{table:resnet50-vs}
\end{table}

\textbf{Varying block size} : In this experiment, we would like to study how accuracy varies with respect to block size. Other parameters like sparsity and number of levels are kept same. From Table \ref{table:resnet50-onelevel} and \ref{table:resnet50-vbs}, we can see that accuracy decreases with increase in block size. This is due to the fact that as we increase block size, more number of high magnitude elements are pruned away due to increased structural constraint. 

\begin{table}[h!]
\centering
\begin{tabular}{|c|c|c|}
\hline
Block-size  & Top-1 Accuracy & Top-5 Accuracy\\ \hline
4x1 & 75.73 (-0.40) & 92.70 (-0.16) \\ \hline
8x1 & 75.19 (-0.94) & 92.49 (-0.37) \\ \hline
16x1 & 75.03 (-1.10) & 92.36 (-0.50) \\ \hline
32x1 & 74.52 (-1.61) & 91.99 (-0.87) \\ \hline
\end{tabular}
\caption{Varying block size with sparsity set to 50\%.}
\label{table:resnet50-onelevel}
\end{table}

\begin{table}[h!]
\centering
\begin{tabular}{|p{2cm}|p{2cm}|c|c|}
\hline
\multicolumn{2}{|c|}{Block-size/sparsity} & \multicolumn{2}{|c|}{Accuracy} \\ \hline
Level-1 (L1) & Level-2 (L2) & Top-1  & Top-5 \\ \hline
4x1/53  & 1x1/97 & 75.93 (-0.20) & 92.81 (-0.05) \\ \hline
8x1/53  & 1x1/97 & 75.67 (-0.46) & 92.65 (-0.21) \\ \hline
16x1/53  & 1x1/97 & 75.55 (-0.58) & 92.75 (-0.11) \\ \hline
32x1/53  & 1x1/97 & 75.26 (-0.87) & 92.48 (-0.38) \\ \hline
\end{tabular}
\caption{Varying block size with two levels. (Cumulative-Sparsity=50)}
\label{table:resnet50-vbs}
\end{table}

\textbf{Varying sparsity distribution:} In this experiment, we set sparsity to 50\% and distribute it across multiple levels with block sizes ranging from 32x1 to 1x1 in a hierarchical fashion. Each row in Table \ref{table:resnet50} corresponds to a Hierarchical block sparse configuration and we can see that accuracy increases by having more fluidity in the structure imposed on sparsity. Another way of bringing fluidity and retaining structure is through Quasi block sparse configuration, which is a subset of Hierarchical block sparse configuration. In this case, there are two levels where one level has block sparsity and another level has unstructured sparsity with block size 1x1. In Quasi block sparse configuration, block sparsity caters for performance and unstructured sparsity caters for accuracy. From Table \ref{table:resnet50-1x1-limit}, we can see that the accuracy increases with increase in fluidity and with minimal loss to accuracy, significant amount of compute can be made regular. 


\begin{table}[h!]
\centering
\begin{tabular}{|c|c|c|c|c|c|c|}
\hline
\multicolumn{5}{|c|}{Sparsity} & \multicolumn{2}{|c|}{Accuracy} \\ \hline
L1-(32x1) & L2-(16x1) & L3-(8x1) & L4-(4x1) & L5-(1x1) & Top-1  & Top-5  \\ \hline
50&-&-&-&-& 74.52 (-1.61) & 91.99 (-0.87)  \\ \hline
75&75&-&-&-& 74.85 (-1.28) & 92.37 (-0.49) \\ \hline
75&87.50&87.50&-&-& 75.18 (-0.95) & 92.49 (-0.37) \\ \hline
75&-&75&-&-& 75.22 (-0.91) & 92.43 (-0.43) \\ \hline
75&87.50&93.75&93.75&-& 75.31 (-0.82) & 92.50 (-0.36) \\ \hline
75&-&-&75&-& 75.44 (-0.69) & 92.63 (-0.23) \\ \hline
75&87.50&93.75&96.875&96.875& 75.63 (-0.50) & 92.67 (-0.19)  \\ \hline
\end{tabular}
\caption{Hierarchical block structure with varying sparsity distribution. (Cumulative-Sparsity=50)}
\label{table:resnet50}
\end{table}

\begin{table}[h!]
\centering
\begin{tabular}{|c|c|c|c||}
\hline
\multicolumn{2}{|c|}{Sparsity} & \multicolumn{2}{|c|}{Accuracy} \\ \hline
L1-(32x1) & L2-(1x1) & Top-1  & Top-5  \\ \hline
62.50 & 87.50 & 75.92 (-0.21) & 92.79 (-0.07) \\ \hline
56.25 & 93.75 & 75.64 (-0.49) & 92.62 (-0.24) \\ \hline
53    & 97   & 75.26 (-0.87) & 92.48 (-0.38) \\ \hline
\end{tabular}
\caption{Quasi block sparsity with varying sparsity distribution. (Cumulative-Sparsity=50)}
\label{table:resnet50-1x1-limit}
\end{table}

\section{Conclusion}
In HBsNN models, levels with smaller block sizes cater for bridging accuracy gap and levels with larger block sizes cater for improving performance. Thus HBsNN models have better accuracies than highly structured sparse models and have better performance than unstructured sparse models. This fluidity in structure in HBsNN models is essential to obtain better sparse models which are both accurate and performant.

\bibliography{iclr2018_conference}

\begin{thebibliography}{10}
\providecommand{\natexlab}[1]{#1}
\providecommand{\url}[1]{\texttt{#1}}
\expandafter\ifx\csname urlstyle\endcsname\relax
  \providecommand{\doi}[1]{doi: #1}\else
  \providecommand{\doi}{doi: \begingroup \urlstyle{rm}\Url}\fi

\bibitem[Cun et~al.(1990)Cun, Denker, and Solla]{Cun90optimalbrain}
Yann~Le Cun, John~S. Denker, and Sara~A. Solla.
\newblock Optimal brain damage.
\newblock In \emph{Advances in Neural Information Processing Systems}, pp.\
  598--605. Morgan Kaufmann, 1990.

\bibitem[Han et~al.(2015)Han, Pool, Tran, and Dally]{han2015learning}
Song Han, Jeff Pool, John Tran, and William Dally.
\newblock Learning both weights and connections for efficient neural network.
\newblock In \emph{Advances in neural information processing systems}, pp.\
  1135--1143, 2015.

\bibitem[Han et~al.(2016)Han, Liu, Mao, Pu, Pedram, Horowitz, and
  Dally]{Han:2016:EEI:3007787.3001163}
Song Han, Xingyu Liu, Huizi Mao, Jing Pu, Ardavan Pedram, Mark~A. Horowitz, and
  William~J. Dally.
\newblock Eie: Efficient inference engine on compressed deep neural network.
\newblock \emph{SIGARCH Comput. Archit. News}, 44\penalty0 (3):\penalty0
  243--254, June 2016.
\newblock ISSN 0163-5964.
\newblock \doi{10.1145/3007787.3001163}.
\newblock URL \url{http://doi.acm.org/10.1145/3007787.3001163}.

\bibitem[Hassibi et~al.(1993)Hassibi, Stork, and Com]{Hassibi93secondorder}
Babak Hassibi, David~G. Stork, and Stork Crc.~Ricoh. Com.
\newblock Second order derivatives for network pruning: Optimal brain surgeon.
\newblock In \emph{Advances in Neural Information Processing Systems 5}, pp.\
  164--171. Morgan Kaufmann, 1993.

\bibitem[He et~al.(2016)He, Zhang, Ren, and Sun]{he2016identity}
Kaiming He, Xiangyu Zhang, Shaoqing Ren, and Jian Sun.
\newblock Identity mappings in deep residual networks.
\newblock In \emph{European Conference on Computer Vision}, pp.\  630--645.
  Springer, 2016.

\bibitem[Narang et~al.(2017{\natexlab{a}})Narang, Diamos, Sengupta, and
  Elsen]{DBLP:journals/corr/NarangDSE17}
Sharan Narang, Gregory~F. Diamos, Shubho Sengupta, and Erich Elsen.
\newblock Exploring sparsity in recurrent neural networks.
\newblock \emph{CoRR}, abs/1704.05119, 2017{\natexlab{a}}.
\newblock URL \url{http://arxiv.org/abs/1704.05119}.

\bibitem[Narang et~al.(2017{\natexlab{b}})Narang, Undersander, and
  Diamos]{DBLP:journals/corr/abs-1711-02782}
Sharan Narang, Eric Undersander, and Gregory~F. Diamos.
\newblock Block-sparse recurrent neural networks.
\newblock \emph{CoRR}, abs/1711.02782, 2017{\natexlab{b}}.
\newblock URL \url{http://arxiv.org/abs/1711.02782}.

\bibitem[Parashar et~al.(2017)Parashar, Rhu, Mukkara, Puglielli, Venkatesan,
  Khailany, Emer, Keckler, and Dally]{DBLP:journals/corr/abs-1708-04485}
Angshuman Parashar, Minsoo Rhu, Anurag Mukkara, Antonio Puglielli, Rangharajan
  Venkatesan, Brucek Khailany, Joel~S. Emer, Stephen~W. Keckler, and William~J.
  Dally.
\newblock {SCNN:} an accelerator for compressed-sparse convolutional neural
  networks.
\newblock \emph{CoRR}, abs/1708.04485, 2017.
\newblock URL \url{http://arxiv.org/abs/1708.04485}.

\bibitem[Szegedy et~al.(2016)Szegedy, Vanhoucke, Ioffe, Shlens, and
  Wojna]{szegedy2016rethinking}
Christian Szegedy, Vincent Vanhoucke, Sergey Ioffe, Jon Shlens, and Zbigniew
  Wojna.
\newblock Rethinking the inception architecture for computer vision.
\newblock In \emph{Proceedings of the IEEE Conference on Computer Vision and
  Pattern Recognition}, pp.\  2818--2826, 2016.

\bibitem[Wen et~al.(2016)Wen, Wu, Wang, Chen, and
  Li]{Wen:2016:LSS:3157096.3157329}
Wei Wen, Chunpeng Wu, Yandan Wang, Yiran Chen, and Hai Li.
\newblock Learning structured sparsity in deep neural networks.
\newblock In \emph{Proceedings of the 30th International Conference on Neural
  Information Processing Systems}, NIPS'16, pp.\  2082--2090, USA, 2016. Curran
  Associates Inc.
\newblock ISBN 978-1-5108-3881-9.
\newblock URL \url{http://dl.acm.org/citation.cfm?id=3157096.3157329}.

\end{thebibliography}
\bibliographystyle{iclr2018_conference}

\end{document}